\documentclass[letterpaper, 10 pt, conference]{ieeeconf}  
\usepackage{amssymb}
\usepackage{amsmath}
\usepackage{graphicx}
\usepackage{float} 

\IEEEoverridecommandlockouts                              

\title{
\LARGE \bf
Dynamics-Aligned Representation Learning under Induced Partial Observability
}

\author{
SM Mazharul Islam$^{\dagger~1}$ and Manfred Huber$^{1}$
\thanks{$^{\dagger}$ corresponding author, {\tt\small mazharul2752@gmail.com}}
\thanks{$^{1}$ are with the Department of Computer Science and Engineering, University of Texas at Arlington, TX-76019, USA}
}

\begin{document}

\maketitle
\thispagestyle{empty}
\pagestyle{empty}

\begin{abstract}
\label{sec:abs}
Model-based reinforcement learning (MBRL) has enabled sample-efficient learning of continuous control policies from high-dimensional observations through learned latent dynamics models. These models are typically trained with observation reconstruction objectives, which provide strong supervision but can bias representations toward visual fidelity leading to an undertrained prior and unreliable long-horizon predictions. We propose a stochastic gating mechanism that induces controlled partial observability during training by sampling a spatial binary mask from a Beta--Bernoulli random field and applying it to the raw observation. The gate probability is resampled at each time step, exposing the encoder to conditions ranging from near-total occlusion to near-full visibility. Integration of our method requires minimal architectural changes from existing dynamics models: the gated observation and the binary mask are channel-wise concatenated and passed to the encoder as a single augmented input, with no modifications to the broader training pipeline. We evaluate the proposed approach on a set of pixel-based continuous control tasks selected from the DeepMind Control Suite. Using model predictive control for evaluation, which relies solely on open-loop model rollouts, we show that stochastic gating improves planning performance and data efficiency across tasks. Further analysis demonstrates that performance is robust to the spatial scale of the mask, and a balanced observability during training outperforms configurations biased toward either sparser or denser observations.
\end{abstract}

\section{INTRODUCTION}
\label{sec:intro}
Recent advances in model-based reinforcement learning (MBRL) have made many control tasks involving high-dimensional visual observations tractable, largely due to the improved sample efficiency offered by compact latent state-space models~\cite{ha2018world, hafner2019learning, wu2023daydreamer}. These approaches learn a latent dynamics model that predicts future states conditioned on the current latent state and action, enabling planning or policy learning in the latent space. Visual observations provide a rich supervision signal for training such models, often through reconstruction-based objectives that encourage the latent state to capture sufficient information about the environment.

However, accurate reconstruction of observations is not the primary objective for control. Instead, the goal is to learn representations that capture the dynamics relevant for decision-making, i.e., how the environment evolves under different action sequences. In many settings, it is sufficient to predict reward trajectories—or even their relative ordering—rather than reconstruct pixel-level observations. This suggests that an ideal representation should be aligned with the underlying dynamics and task-relevant signals, rather than the full observation distribution.

Consequently, learning an accurate prior over latent states is critical, as it governs multi-step predictions used in planning and policy optimization. Previous works \cite{ma2023harmonydream, zhao2017infovae} have shown that strong reconstruction objectives combined with expressive decoders can lead to representations that overfit to observations while under-utilizing the learned dynamics. In such cases, the latent dynamics model may fail to generalize under open-loop rollouts, limiting its usefulness for downstream control.

Although most MBRL methods are formulated under the partially observable Markov decision process (POMDP) framework~\cite{kaelbling1998planning}, many commonly used benchmarks are effectively fully observable for large portions of the state space. For example, tasks in the DeepMind Control Suite \cite{tassa2018deepmind} used in works such as PlaNet \cite{hafner2019learning} and Dreamer series \cite{hafner2019dream, hafner2023mastering} provide observations that are often sufficient to infer the underlying state with little ambiguity.

In such settings, posterior inference becomes relatively easy, as the model can rely heavily on the current observation. In contrast, learning a consistent and predictive prior remains challenging. This imbalance, combined with powerful decoders, often leads to a failure mode where the model ignores the prior during training \cite{bowman2016generating}, wherein the KL divergence between posterior and prior is driven toward zero and the dynamics model receives no meaningful gradient signal to develop predictive representations. As a result, the learned dynamics model becomes unreliable for open-loop prediction, which is precisely the regime required for planning and policy optimization.

\begin{figure*}[ht]
      \centering
      \includegraphics[width=0.95\linewidth]{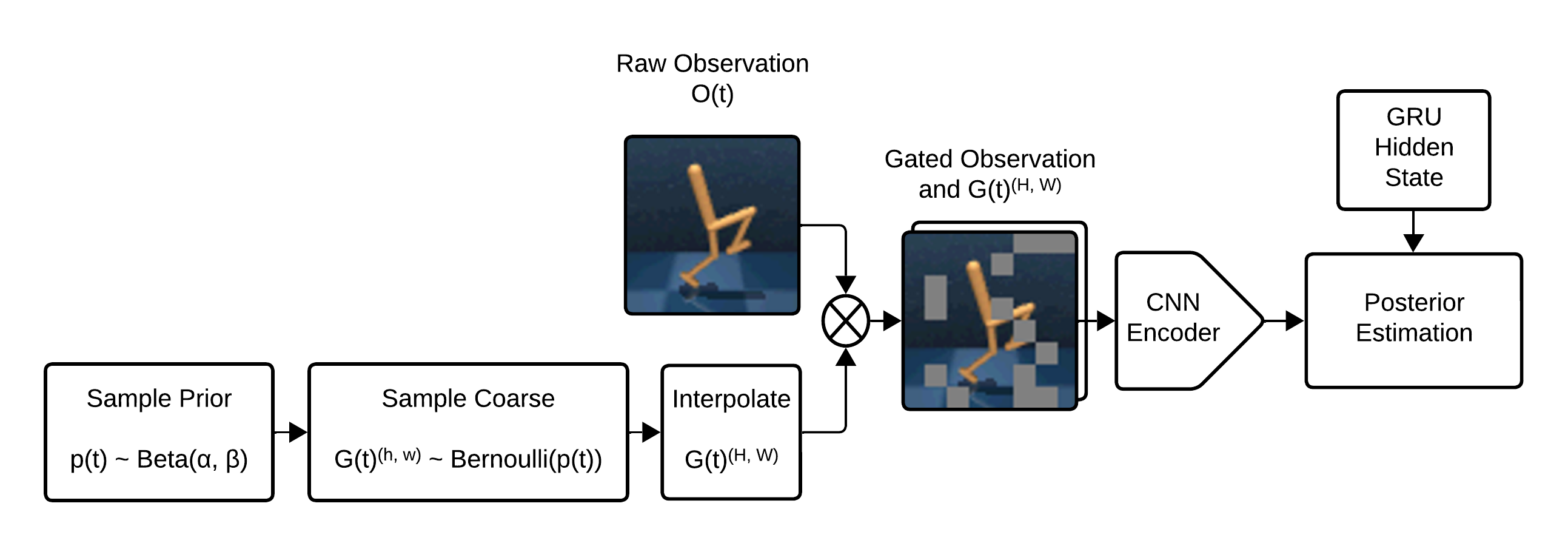}
      \caption{Posterior estimation under induced partial observability. By limiting observation access, stochastic gating prevents posterior collapse onto observations and drives learning in the prior.}
      \label{fig:overview}
   \end{figure*}

To address both the decoder dominance due to trivial posterior estimation and as a consequence collapsed posterior, we propose a stochastic gating mechanism that enforces partial observability conditions even in environments where such conditions are only loosely satisfied. The core idea is to randomly mask pixels in the input observation before encoding, simulating the information loss characteristic of a true POMDP. The encoder is modified minimally: a single additional channel is introduced to carry the gate state, indicating which pixels survived the masking process. The partially occluded observation and the gate state are jointly encoded to produce the posterior estimate, preserving the encoder's ability to exploit available information while preventing it from trivially inverting the full observation. Figure~\ref{fig:overview} illustrates an overview of our proposed method.

Since stochastic gating introduces a variable number of active pixels at each timestep, the observation reconstruction loss must be adapted accordingly. We restrict the reconstruction loss to only the pixels that survive the gating process. The values of masked pixels are unobserved and, therefore, do not contribute to a meaningful reconstruction target. This loss-asymmetry creates a strong inductive bias toward routing information through the prior branch, which is precisely the dynamics model that we seek to learn. We stabilize prior learning through KL balancing \cite{hafner2019dream} and free bits \cite{kingma2013auto}.

We evaluate the proposed method on a set of DeepMind Control Suite (DMC) tasks~\cite{tassa2018deepmind} that are near-fully observable. To isolate the effect of the learned representations from confounds introduced by policy optimization, we deliberately avoid the commonly used actor-critic framework~\cite{haarnoja2018soft, schulman2017proximal}. Instead, we evaluate policy quality solely through model-predictive control via cross entropy method (MPC-CEM)~\cite{richards2005robust}, which relies entirely on the model's open-loop predictive capability to select actions. This design choice allows us to attribute performance differences directly to the quality of the learned dynamics model rather than to the policy optimizer. Experimental results demonstrate that partial observation though stochastic gating consistently improves downstream task performance over the baseline, enabling the model to identify better policies in fewer environment steps.

The main contributions of this work are as follows:

\begin{itemize}
    \item We propose a stochastic gating mechanism that induces partial observability at the input level, providing a principled way to prevent posterior dominance and collapse in near-fully observable environments.
    \item We introduce a gating-aware reconstruction loss that conditions supervision on the surviving pixels, coupled with KL balancing and free bits to stabilize prior learning throughout training.
    \item The proposed module integrates seamlessly into existing MBRL architectures based on the Recurrent State Space Model (RSSM)~\cite{hafner2019learning}, requiring only a single additional encoder input channel and no changes to the broader training regimen.
    \item Through MPC-CEM evaluation on DMC tasks, we demonstrate that the proposed approach remains robust under the spatial scale of gating and improves task performance when an agent is trained with a balanced visibility condition.
\end{itemize}

The rest of this paper develops and evaluates our approach to address posterior collapse and facilitate dynamics-aligned representation learning in MBRL. Section \ref{sec:rel_work} surveys relevant previous work on latent dynamics modeling and representation learning with partial observability in MBRL. Section \ref{sec:prelim} introduces the necessary background on MBRL, including the baseline RSSM method. In Section \ref{sec:method}, we formalize our stochastic gating method, related loss terms, and it's integration with RSSM. Section \ref{sec:exp} presents experimental results with task performance on continuous control tasks and ablation studies evaluating the effect of key hyperparameters. Finally, Section \ref{sec:concl} concludes the paper and discusses potential directions for future research.

\section{RELATED WORK}
\label{sec:rel_work}

\subsection{Latent Dynamics Modeling in MBRL}
Model-based reinforcement learning (MBRL) methods \cite{ha2018world, watter2015embed, zhang2019solar} learn compact latent representations of environment dynamics to enable sample-efficient planning and policy evaluation. Early approaches such as PlaNet \cite{hafner2019learning} and Dreamer \cite{hafner2019dream} use recurrent state-space models (RSSMs) to encode high-dimensional image observations into latent states and predict future transitions. These methods formalize the control problem as a POMDP, where the RSSM maintains a belief state over the hidden environment state conditioned on the history of observations and actions. These methods are typically trained using a variational objectives \cite{kingma2013auto} where observation reconstruction provides the dominating supervised signal. However, this objective can bias the representation toward modeling high-dimensional visual details that are not directly relevant for control \cite{ma2023harmonydream, zhao2017infovae}. As a result, the learned latent dynamics may be less effective for long-horizon prediction and decision-making, particularly when used in open-loop rollouts.

Alternative architectures address this tension differently. TD-MPC \cite{hansen2022temporal} learns a task-oriented latent dynamics model together with a terminal value function via temporal difference learning, using short-horizon MPC for planning rather than full observation reconstruction. IRIS \cite{micheli2022transformers} takes a complementary approach, casting dynamics learning as a sequence modeling problem by pairing a discrete autoencoder with an autoregressive Transformer. More recent work such as HRSSM \cite{sun2024learning} applies spatio-temporal masking and bisimulation-based objectives within an RSSM framework to suppress irrelevant exogenous observations and improve task-relevant representation learning. These lines of work collectively highlight that observation reconstruction, while providing a dense training signal, is neither necessary nor always sufficient to learn reward-predictive representations.

\subsection{Partial Observability in Reinforcement Learning}
Partial observability is a fundamental challenge in sequential decision-making. In a typical POMDP setting \cite{kaelbling1998planning}, an agent cannot directly observe the underlying environment state and must infer it from a history of incomplete observations. Hence, POMDPs are a general framework for sequential decision-making under latent state uncertainty, yet learning in POMDPs is intractable in the worst case. A classical approach to tractable POMDP learning is to maintain a belief state, i.e. a distribution over hidden states, and plan with respect to it \cite{cassandra1994acting}. In deep RL, recurrent architectures such as LSTMs and GRUs serve as implicit belief state estimators. Heess et al. \cite{heess2015memory} demonstrate that extending deterministic and stochastic policy gradient methods with LSTM-based recurrence enables agents to solve a range of continuous control problems with partial observability, including tasks that require integration of noisy sensor information and identification of hidden system parameters.

In the MBRL setting, partial observability arises naturally since image observations often do not reveal the full environment state. The RSSM \cite{hafner2019learning} addresses this by separating the latent state into deterministic and stochastic components, where the stochastic component functions as a compact belief over the current hidden state. A distinct but related line of work considers structural forms of partial observability, such as observation delays \cite{chen2011robust} and missing sensor modalities. Masked World Models \cite{seo2023masked} decouple visual representation learning from dynamics learning by training an autoencoder to reconstruct pixels from masked convolutional features, introducing an auxiliary reward prediction objective to encode task-relevant information.

\section{PRELIMINARIES}
\label{sec:prelim}

\subsection{Model Based Reinforcement Learning}
We formulate the problem of learning RL tasks from high-dimensional visual observations as a partially observable Markov decision process (POMDP), which is defined as a tuple $(\mathcal{O}, \mathcal{A}, p, r, \gamma)$. $\mathcal{O}$ is the visual observation space and $\mathcal{A}$ represents the action space. At timestep $t$, an agent generates an action $a_t \sim \pi$ and receives the observation $o_{t} \sim p(o_{t} | o_{<t}, a_{<t})$ and scaler reward signal $r_t \sim r(o_{\leq t}, a_{<t})$ according to the unknown environment dynamics $p$ and reward function $r$. Through the approximation of the underlying environment $(p, r)$, the goal of an MBRL agent is to maximize the $\gamma$-discounted cumulative return $\mathbb{E}\left[ \sum_{t-1}^T \gamma^{t-1} r_t\right]$.

\subsection{Recurrent State Space Model (RSSM)}
The Recurrent State Space Model (RSSM) \cite{hafner2019learning} is a latent dynamics model that learns a compact representation of the environment from sequences of observations, actions, and rewards $\{o_t, a_t, r_t\}^{T}_{t=1}$ sampled from a replay buffer. It jointly learns four components:

\begin{equation}
    \begin{aligned}
        \text{Representation model: } & s_t \sim q_{\theta}(s_t \mid s_{t-1}, a_{t-1}, o_t) \\
        \text{Observation model: } & \hat{o}_t \sim p_{\theta} (\hat{o}_t \mid s_t) \\
        \text{Transition model: } &  \hat{s}_t \sim p_{\theta}(\hat{s}_t \mid s_{t-1}, a_{t-1}) \\
        \text{Reward model: } & \hat{r}_t \sim   p_{\theta}(\hat{r}_t \mid s_t).
    \end{aligned}
    \label{rssm}
\end{equation}

The representation model estimates the posterior of the latent state $s_t$ from the previous state $s_{t-1}$, the previous action $a_{t-1}$, and the current observation $o_t$. The transition model estimates the prior using only $s_{t-1}$ and $a_{t-1}$ without access to $o_t$. Together they define a nonlinear Kalman filtering framework: the representation model refines the state estimate when observations are available, while the transition model propagates belief in their absence. Under our partial observation scheme (Section~\ref{sec:gated_obs}), $o_t$ is replaced by $\mathrm{Concat}(\mathbf{G}(t) \odot o_t, \mathbf{G}(t))$, meaning the representation model receives only a stochastic subset of the complete observed pixels at each training step. All parameters $\theta$ are trained by jointly minimizing:
\begin{equation}
    \begin{aligned}
        \text{Observation loss: } & \mathcal{L}_o(\theta) = -\log p_{\theta}(o_t \mid s_t) \\
        \text{Reward loss: } & \mathcal{L}_r(\theta) = -\log p_{\theta}(r_t \mid s_t) \\
        \text{Dynamics loss: } &  \mathcal{L}_d(\theta) = \mathrm{KL}\big(q_{\theta}(s_t \mid s_{t-1}, a_{t-1}, o_t) \\
        & \qquad\qquad\quad \|\, p_{\theta}(s_t \mid s_{t-1}, a_{t-1})\big),
    \end{aligned}
    \label{eq:rssm_loss}
\end{equation}

The dynamics loss minimizes the KL divergence between the posterior and the prior, encouraging the transition model to remain a faithful one-step predictor of the latent state. Under the standard assumption that the observation and reward models are parameterized as Gaussian distributions, $\mathcal{L}_o$ and $\mathcal{L}_r$ reduce to mean-squared error between predictions and targets. The observation loss $\mathcal{L}_o(\theta)$ is replaced by the gated variant $\mathcal{L}_o^G(\theta)$ (Section~\ref{sec:gated_loss}) when training under partial observability.

During training, the replay buffer is initialized with a set of seed episodes collected under random actions, and subsequent data is gathered by perturbing the optimal action with small Gaussian noise for exploration purpose. To reduce the effective planning horizon and stabilize learning, each action is repeated $R$ times. Using model predictive control via cross-entropy method (MPC-CEM)~\cite{rubinstein1997optimization,chua2018deep}, at each planning step candidate sequences are sampled from a fctorized belief over action sequences, evaluated under the learned world model, and then the distribution is refit to the top-$K$ candidates for $I$ iterations.

\section{INDUCED PARTIAL OBSERVABILITY}
\label{sec:method}

\subsection{Partial Observation}
\label{sec:gated_obs}
We introduce a time-dependent partial observation mechanism based on a Random Bernoulli Field defined over a 2D spatial lattice. At each time step $t$, we first sample a gating probability:
\begin{equation}
    \begin{aligned}
        p(t) \sim \mathrm{Beta}(\alpha, \beta), \quad \alpha, \beta > 0,
    \end{aligned}
    \label{eq:gating_0}
\end{equation}

This governs the expected density of observed locations in the field. Then Conditioned on $p(t)$, we generate a binary mask over a low-resolution spatial grid:
\begin{equation}
    \begin{aligned}
        \mathbf{G}(t)^{\mathrm{coarse}} &\sim \mathrm{Bernoulli}(p(t)), \\
        \mathbf{G}(t)^{\mathrm{coarse}} &\in \{0, 1\}^{h \times w}.
    \end{aligned}
    \label{eq:gating_1}
\end{equation}

This coarse mask is up-sampled to the full observation resolution via nearest-neighbor interpolation:
\begin{equation}
    \begin{aligned}
        \mathbf{G}(t) = \mathrm{Up}(\mathbf{G}(t)^{\mathrm{coarse}}) \in \{0, 1\}^{H \times W},
    \end{aligned}
    \label{eq:gating_2}
\end{equation}

where $H \times W$ denotes the spatial resolution of the observation. Under the isotropic assumption, a single scale factor $s$ governs both spatial dimensions, where
\begin{equation*}
    \begin{aligned}
        s = H/h = W/w.
    \end{aligned}
    \label{eq:gating_3}
\end{equation*}

The final partial observation is obtained by concatenating the gated observation $\mathbf{G}(t) \odot o_t$ and the binary gate $\mathbf{G}(t)$ along the channel dimension:
\begin{equation}
    \begin{aligned}
        \tilde{o}_t = \mathrm{Concat}(\mathbf{G}(t) \odot o_t, \mathbf{G}(t)),
    \end{aligned}
    \label{eq:gating_4}
\end{equation}
where $\odot$ denotes the Hadamard (element-wise) product.

\subsection{Stochastic Gating}

A fixed gating probability (i.e., constant $p$) restricts the model to a narrow regime of observability, which can lead to brittle representations that do not generalize across different levels of information availability. In contrast, sampling $p(t)$ at each time step exposes the encoder to a continuum of observability conditions, ranging from near-complete to highly partial observations. This variability encourages the model to learn representations that are robust to missing information while still retaining useful features when observations are fully available.

At evaluation time, we set $\mathbf{G}(t) = 1$, i.e., no masking is applied. Because the model has been trained across varying levels of observability, it can leverage full observations during testing without being over-reliant on them while training.

\subsection{Significance of Beta Prior}
A natural baseline is to sample the gating-prior from a uniform distribution, i.e. $p(t) \sim \mathrm{Uniform}(0, 1)$. However, this corresponds to a special case of the Beta distribution:
\begin{equation*}
    \begin{aligned}
        \mathrm{Uniform}(0, 1) \equiv \mathrm{Beta}(1, 1)
    \end{aligned}
    \label{eq:gating_5}
\end{equation*}

Hence, considering its flexibility in shaping the distribution over gating probabilities, we opt for Beta distribution to model the gating-prior. By adjusting $(\alpha, \beta)$, we can control the frequency of low-visibility versus high-visibility regimes. For example, $\alpha > \beta$ will generate gating-mask biased towards high visibility while $\alpha < \beta$ will bias the visibility towards low. This flexibility allows us to tailor the training distribution to emphasize challenging regimes (e.g., sparse observations) while still occasionally exposing the model to near-complete inputs.

\subsection{Gated Observation Loss}
\label{sec:gated_loss}
During training, the encoder observes only a stochastic subset of the full image, determined by the time-varying gating mask $\mathbf{G}(t)$ as defined by Equation~\eqref{eq:gating_0}--\eqref{eq:gating_2}. Penalizing the decoder for reconstruction error over unobserved pixels would be ill-posed, as the posterior state $s_t$ carries no information about those locations. To address this, we gate the per-pixel log-likelihood by $\mathbf{G}(t)$ before aggregation, restricting the reconstruction signal to observed pixels only:

\begin{equation}
    \mathcal{L}_o(\theta) = 
    - \sum_{i, j} \mathbf{G}(t)^{ij} \log p_{\theta}(o^{ij}_t \mid s_t)
\label{eq:gated_obs_loss}
\end{equation}

where $o_t^{ij}$ denotes the pixel at location $(i, j)$ of complete observation $o_t$, and $\mathbf{G}(t)^{ij}$ is the corresponding gating value at full resolution. This replaces the observation loss term in Equation~\ref{eq:rssm_loss}.

\subsection{KL Balancing}
\label{sec:kl_balance}
The original RSSM uses \emph{free bits} \cite{kingma2013auto} to prevent premature posterior collapse. Later work \cite{hafner2020mastering} introduces \emph{KL balancing}, which instead controls gradient flow between the posterior and prior by decomposing the KL into two asymmetric terms with selective gradient stopping:

\begin{equation}
    \begin{aligned}
        \mathcal{L}_d =\;& 
        \alpha \cdot \mathrm{KL}\big(\mathrm{sg}[q_{\theta}(s_t \mid \cdot)] \,\|\, p_{\theta}(s_t \mid \cdot)\big) \\
        &+ (1 - \alpha) \cdot \mathrm{KL}\big(q_{\theta}(s_t \mid \cdot) \,\|\, \mathrm{sg}[p_{\theta}(s_t \mid \cdot)]\big),
    \end{aligned}
\end{equation}

where $\mathrm{sg}[\cdot]$ denotes the stop-gradient operator and $\alpha \in [0,1]$ controls the relative learning rates of the prior and posterior. Under stochastic partial observability, KL balancing becomes essential. The gating mask $\mathbf{G}(t)$ reduces the information available to the encoder, weakening the posterior and lowering the KL term in expectation. Choosing $\alpha > 0.5$ shifts learning toward the prior, encouraging it to better model the underlying dynamics. Following \cite{hafner2020mastering}, we set $\alpha = 0.8$.

\section{EXPERIMENTS}
\label{sec:exp}
We evaluate the proposed method against variants trained under full observability on four continuous control tasks from the DeepMind Control Suite \cite{tassa2018deepmind}: Cartpole Swingup, Cheetah Run, Reacher Easy, and Walker Walk. All tasks use pixel observations of size $3 \times 64 \times 64$. Following \cite{hafner2019learning}, we set the action repeat to 8 for Cartpole, 4 for Cheetah and Reacher, and 2 for Walker, and adopt the RSSM as the baseline dynamics model and use the original training schedule and hyperparameters, with two modifications: (i) latent overshooting is removed, and (ii) KL balancing \cite{hafner2020mastering} with $\alpha = 0.8$ is applied to stabilize prior learning (Section~\ref{sec:kl_balance}).

\begin{figure}[ht]
      \centering
      \includegraphics[width=0.99\linewidth]{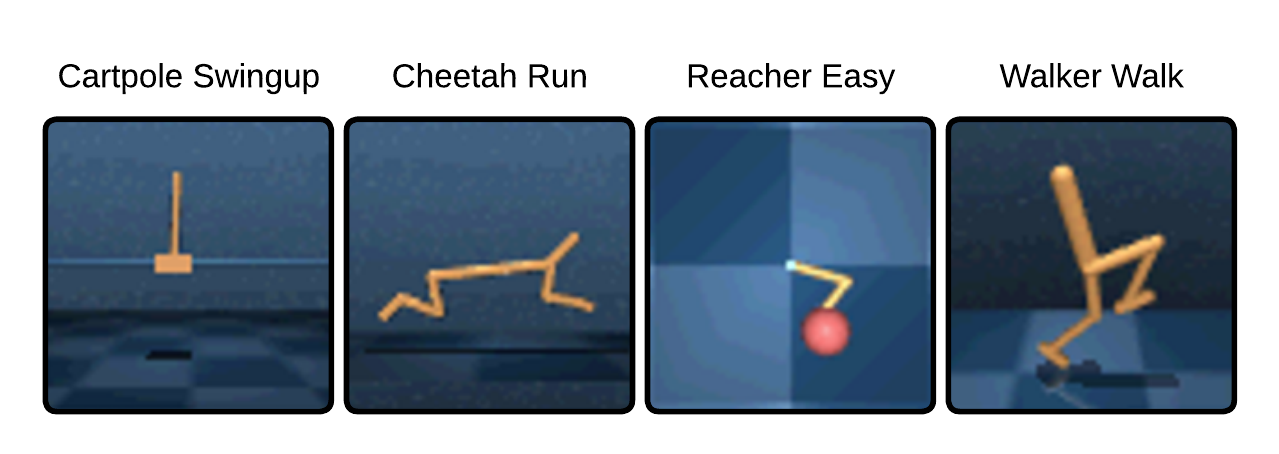}
      \caption{Evaluation environments from the DeepMind Control Suite.}
      \label{fig:env}
\end{figure}

Specifically, we address three questions: (1) whether training under partial observability improves policy learning over full observability; (2) sensitivity to the spatial scale factor $s$ of the gating field; and (3) the effect of the Beta prior parameters $(\alpha, \beta)$ on learned representations. All experiments use CEM-MPC for action selection based on open-loop model rollouts, isolating the effect of the learned dynamics from the policy optimization procedure.

\subsection{Induced Partial Observation vs Raw Observation}
\begin{figure}[h]
      \centering
      \includegraphics[width=0.99\linewidth]{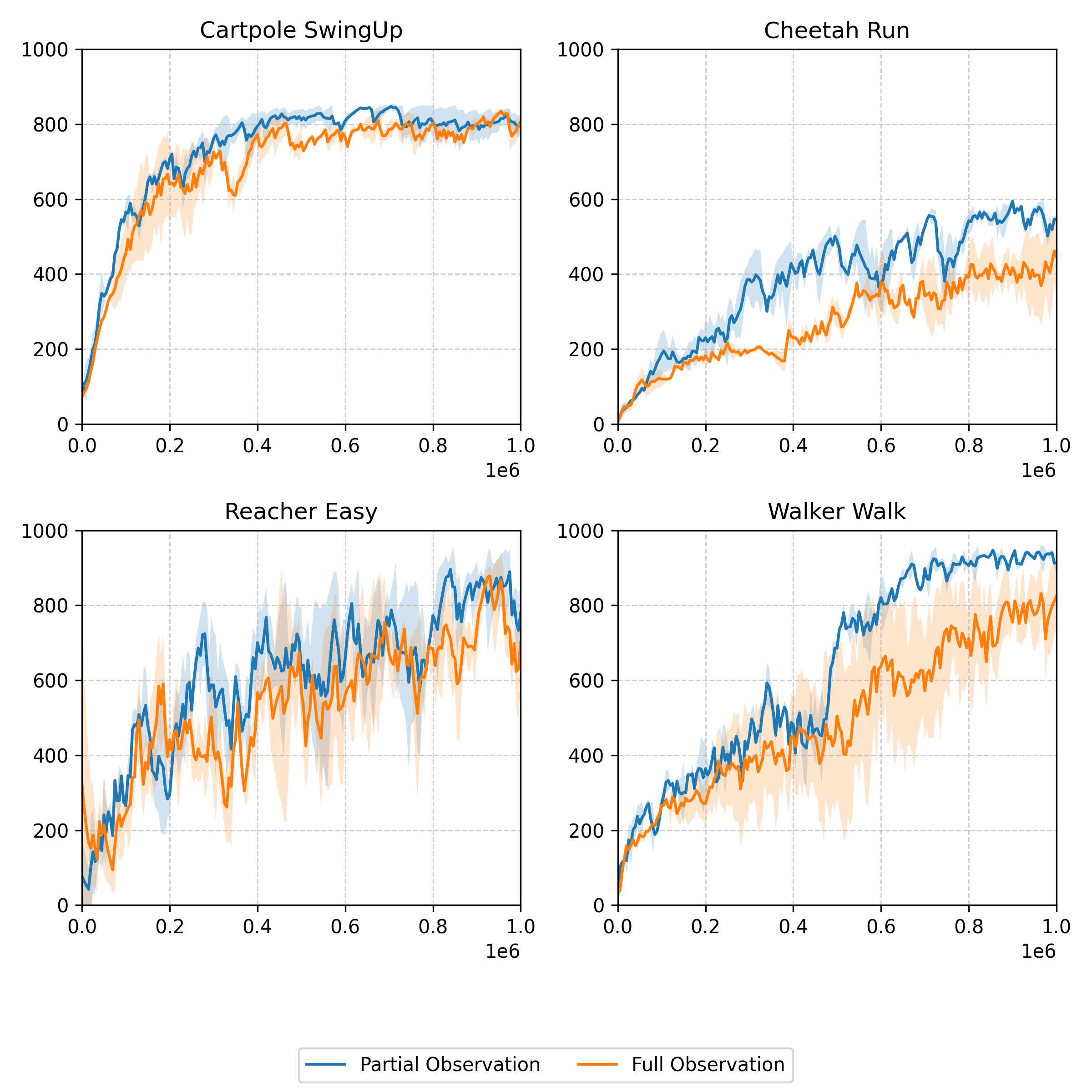}
      \caption{Comparison of task learning performance under induced partial observation and full observation. Each subplot shows test return versus environment steps. Stochastic gating improves learning efficiency and final performance on Cheetah, Reacher, and Walker, while matching on Cartpole. The lines show the average and the shaded area represent 1 standard deviation over 5 seeds.}
      \label{fig:exp_2}
\end{figure}

Figure \ref{fig:exp_2} compares episodic returns under induced partial observability and full-observation training, using identical training and planning settings. Cartpole shows nearly identical performance across both methods, likely because the task is already partially observable—the pendulum can leave the camera view at extreme angles, limiting the additional impact of gating.

In contrast, Cheetah Run, Reacher, and Walker consistently outperform the full-observation baseline under induced partial observability. These improvements indicate that stochastic observation masking promotes dynamics-consistent latent representations, leading to better multi-step open-loop predictions under model predictive control.

\subsection{Effect of Spatial Scale}
Next, we ablate the spatial scale factor $s = H/h = W/w$, which controls the granularity of the masked regions. Larger values of $s$ produce finer block-structured masks with smaller contiguous occlusions, while $s = 1$ corresponds to independent dropout at the observation-level. We experiment with $s \in {1, 2, 4}$, while keeping the Beta prior fixed at $\text{Beta}(2, 2)$ across all conditions so that the expected observation density $\mathbb{E}[p(t)] = 0.5$ remains constant. This setup isolates the effect of spatial structure from the overall level of observability.

\begin{figure}[h]
    \centering
    \includegraphics[width=0.9\linewidth]{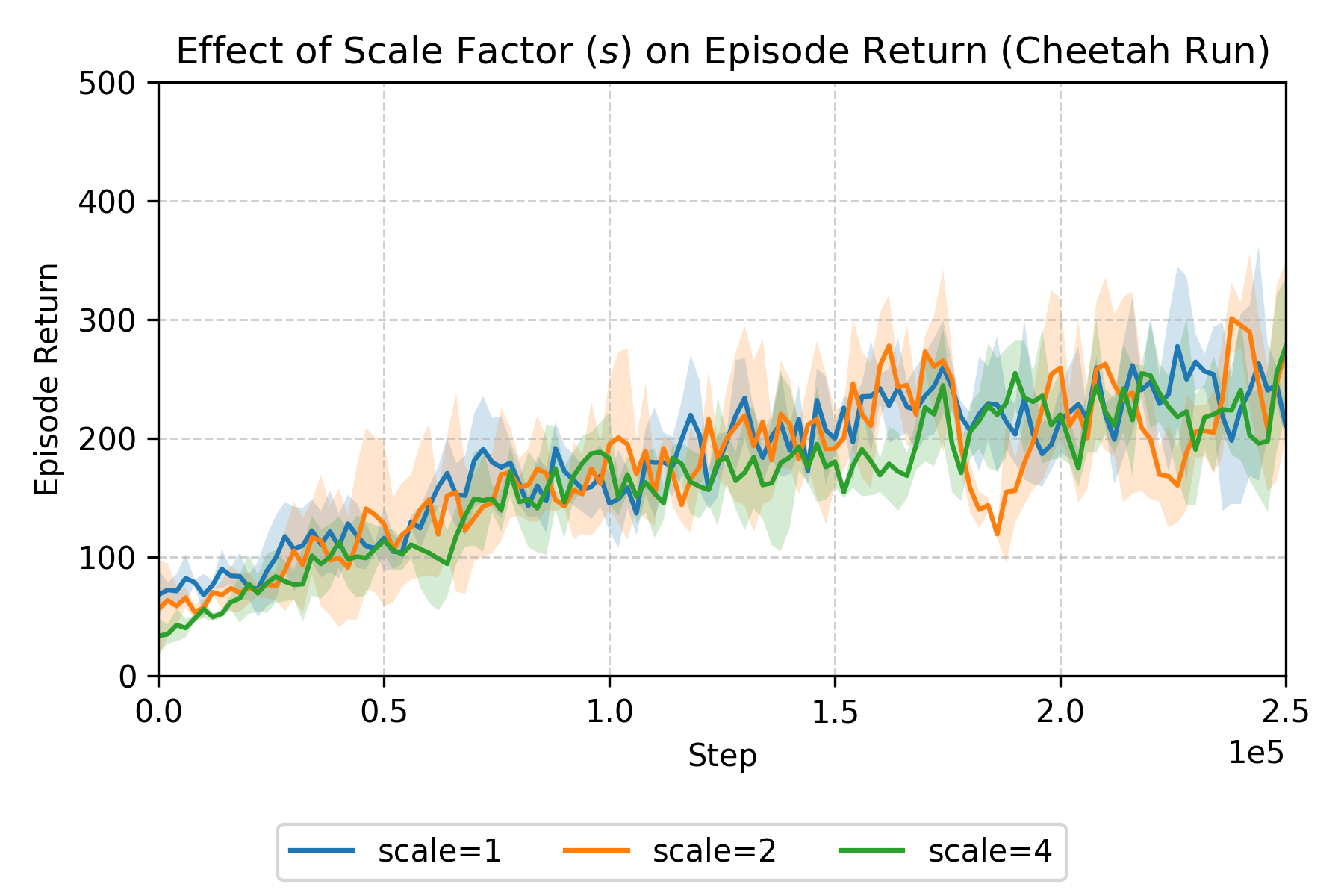}
    \caption{Comparison of scale factors with fixed $\text{Beta}(2, 2)$. Performance curves are comparable across different scales, indicating that task performance is invariant to the spatial granularity of the mask when the expected observation density is constant. The lines show the average and the shaded areas represent 1 standard deviation over 5 seeds.}
    \label{fig:exp_0}
\end{figure}

One might expect that coarser masks, which occlude semantically coherent spatial regions, would pose a more challenging inference problem than fine-grained dropout, and thus affect dynamics learning differently. However, we find that task performance is largely invariant to the choice of $s$, as shown in Figure~\ref{fig:exp_0}, where across all three scale factors, the test performances converge to comparable episode returns. This suggests that the degree of partial observability, as controlled by the Beta prior, plays a more dominant role in shaping dynamics learning than the spatial structure of the masking.

\subsection{Effect of Beta Prior Parameters}
Finally, we investigate the effect of the Beta prior shape parameters $(\alpha, \beta)$ by evaluating three configurations that differ in both expected observation density and concentration. The distribution $\text{Beta}(2, 8)$ places most of its mass near $p(t) \approx 0$, biasing training toward sparse observations in which most pixels are masked. In contrast, $\text{Beta}(8, 2)$ concentrates mass near $p(t) \approx 1$, resulting in predominantly dense observations that are close to full visibility. The symmetric $\text{Beta}(2, 2)$ has mean $\mathbb{E}[p(t)] = 0.5$ and moderate variance, thereby exposing the encoder to a broad range of observability levels throughout training. The spatial scale is fixed at $s = 2$ for all configurations.

\begin{figure}[h]
    \centering
    \includegraphics[width=0.9\linewidth]{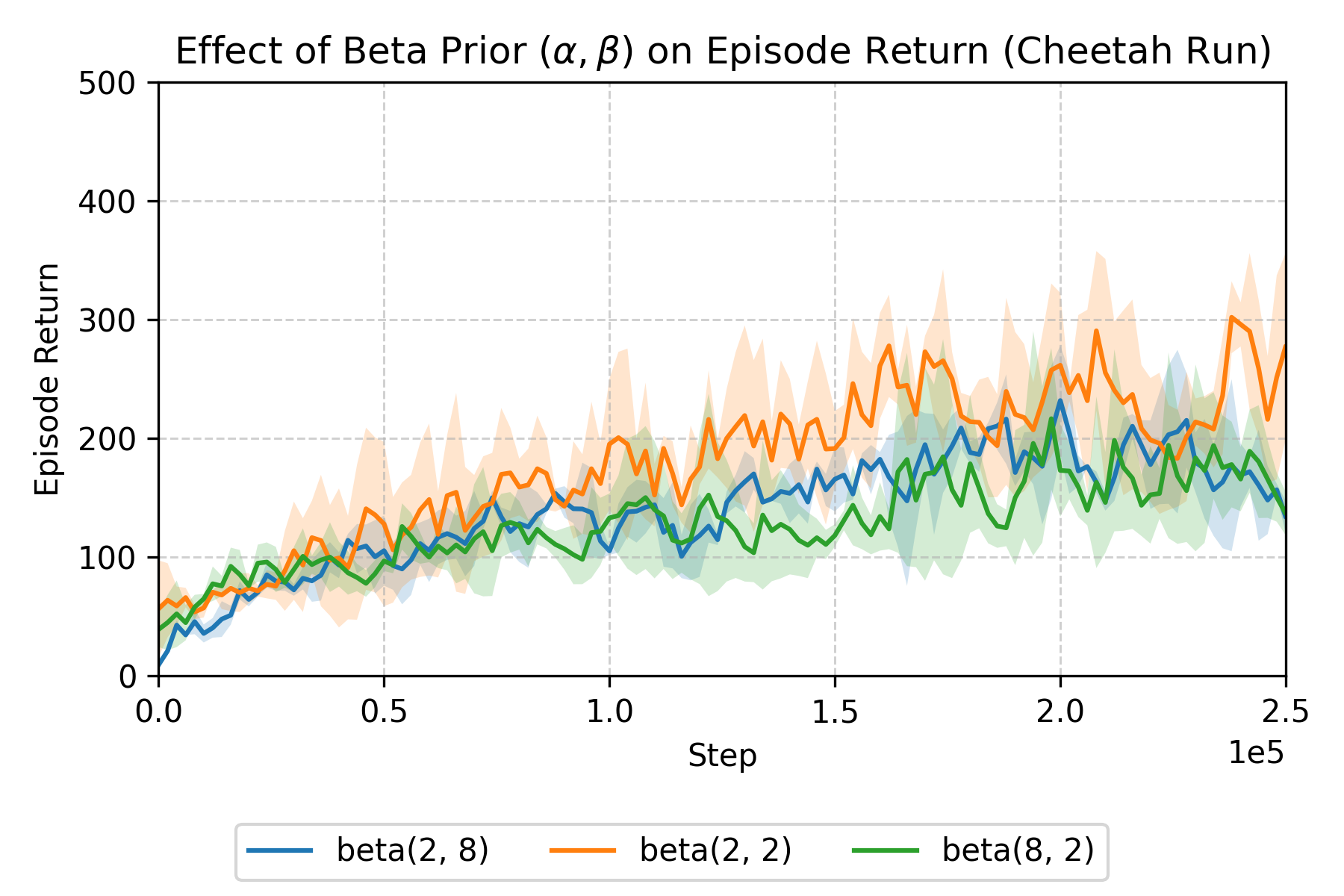}
    \caption{Comparison of Beta prior configurations. The plot shows test performance over environment steps. The symmetric prior consistently outperforms the asymmetric alternatives, indicating that balanced exposure across the full range of observability during training is more critical than any specific directional bias in observability. The lines show the average and the shaded area represent 1 standard deviation over 5 seeds.}
    \label{fig:exp_1}
\end{figure}

As shown in Figure~\ref{fig:exp_1}, $\text{Beta}(2, 2)$ outperforms both asymmetric alternatives by a comparable margin. This symmetric outcome indicates that the observed performance gains are not driven by any particular level of partial observability. Instead, they arise from training under a diverse range of observability conditions, which appears to promote more effective representation learning.

\section{CONCLUSION}
\label{sec:concl}
In this work, we identify over-reliance on instantaneous observations as a key limitation in model-based reinforcement learning, leading to latent representations that are weakly aligned with underlying dynamics. To address this, we introduce a stochastic gating mechanism based on a Beta–Bernoulli process that induces controlled partial observability during training, effectively serving as an information bottleneck. Empirical results on tasks from the DeepMind Control Suite show that training under partial observability improves planning performance and data efficiency. Performance remains stable across different spatial masking scales, and training under diverse observability conditions proves more effective than biasing toward either sparse or dense observations alone. These results indicate that exposure to a broad range of information regimes promotes more dynamics-aligned representations. Future work includes extending evaluation to more complex environments and integrating the approach with actor–critic methods to assess its impact on end-to-end policy learning.







\bibliographystyle{IEEEtran}
\bibliography{ref}

@inproceedings{cassandra1994acting,
  title={Acting optimally in partially observable stochastic domains},
  author={Cassandra, Anthony R and Kaelbling, Leslie Pack and Littman, Michael L},
  booktitle={Aaai},
  volume={94},
  pages={1023--1028},
  year={1994}
}

@article{hafner2019dream,
  title={Dream to control: Learning behaviors by latent imagination},
  author={Hafner, Danijar and Lillicrap, Timothy and Ba, Jimmy and Norouzi, Mohammad},
  journal={arXiv preprint arXiv:1912.01603},
  year={2019}
}

@article{ha2018world,
  title={World models},
  author={Ha, David and Schmidhuber, J{\"u}rgen},
  journal={arXiv preprint arXiv:1803.10122},
  year={2018}
}

@article{hafner2020mastering,
  title={Mastering atari with discrete world models},
  author={Hafner, Danijar and Lillicrap, Timothy and Norouzi, Mohammad and Ba, Jimmy},
  journal={arXiv preprint arXiv:2010.02193},
  year={2020}
}

@phdthesis{richards2005robust,
  title={Robust constrained model predictive control},
  author={Richards, Arthur George},
  year={2005},
  school={Massachusetts Institute of Technology}
}

@inproceedings{bowman2016generating,
  title={Generating sentences from a continuous space},
  author={Bowman, Samuel and Vilnis, Luke and Vinyals, Oriol and Dai, Andrew and Jozefowicz, Rafal and Bengio, Samy},
  booktitle={Proceedings of the 20th SIGNLL conference on computational natural language learning},
  pages={10--21},
  year={2016}
}

@article{chen2011robust,
  title={Robust Kalman filtering for uncertain state delay systems with random observation delays and missing measurements},
  author={Chen, B and Yu, L and Zhang, W-A},
  journal={IET Control Theory \& Applications},
  volume={5},
  number={17},
  pages={1945--1954},
  year={2011},
  publisher={IET}
}

@inproceedings{seo2023masked,
  title={Masked world models for visual control},
  author={Seo, Younggyo and Hafner, Danijar and Liu, Hao and Liu, Fangchen and James, Stephen and Lee, Kimin and Abbeel, Pieter},
  booktitle={Conference on Robot Learning},
  pages={1332--1344},
  year={2023},
  organization={PMLR}
}

@article{zhao2017infovae,
  title={Infovae: Information maximizing variational autoencoders},
  author={Zhao, Shengjia and Song, Jiaming and Ermon, Stefano},
  journal={arXiv preprint arXiv:1706.02262},
  year={2017}
}

@inproceedings{wu2023daydreamer,
  title={Daydreamer: World models for physical robot learning},
  author={Wu, Philipp and Escontrela, Alejandro and Hafner, Danijar and Abbeel, Pieter and Goldberg, Ken},
  booktitle={Conference on robot learning},
  pages={2226--2240},
  year={2023},
  organization={PMLR}
}

@article{kingma2013auto,
  title={Auto-encoding variational bayes},
  author={Kingma, Diederik P and Welling, Max},
  journal={arXiv preprint arXiv:1312.6114},
  year={2013}
}

@article{heess2015memory,
  title={Memory-based control with recurrent neural networks},
  author={Heess, Nicolas and Hunt, Jonathan J and Lillicrap, Timothy P and Silver, David},
  journal={arXiv preprint arXiv:1512.04455},
  year={2015}
}

@article{sun2024learning,
  title={Learning latent dynamic robust representations for world models},
  author={Sun, Ruixiang and Zang, Hongyu and Li, Xin and Islam, Riashat},
  journal={arXiv preprint arXiv:2405.06263},
  year={2024}
}

@article{micheli2022transformers,
  title={Transformers are sample-efficient world models},
  author={Micheli, Vincent and Alonso, Eloi and Fleuret, Fran{\c{c}}ois},
  journal={arXiv preprint arXiv:2209.00588},
  year={2022}
}

@article{hansen2022temporal,
  title={Temporal difference learning for model predictive control},
  author={Hansen, Nicklas and Wang, Xiaolong and Su, Hao},
  journal={arXiv preprint arXiv:2203.04955},
  year={2022}
}

@article{kaelbling1998planning,
  title={Planning and acting in partially observable stochastic domains},
  author={Kaelbling, Leslie Pack and Littman, Michael L. and Cassandra, Anthony R.},
  journal={Artificial Intelligence},
  volume={101},
  number={1-2},
  pages={99--134},
  year={1998}
}

@article{chua2018deep,
  title={Deep reinforcement learning in a handful of trials using probabilistic dynamics models},
  author={Chua, Kurtland and Calandra, Roberto and McAllister, Rowan and Levine, Sergey},
  journal={Advances in neural information processing systems},
  volume={31},
  year={2018}
}

@article{rubinstein1997optimization,
  title={Optimization of computer simulation models with rare events},
  author={Rubinstein, Reuven Y},
  journal={European Journal of Operational Research},
  volume={99},
  number={1},
  pages={89--112},
  year={1997},
  publisher={Elsevier}
}

@article{ma2023harmonydream,
  title={Harmonydream: Task harmonization inside world models},
  author={Ma, Haoyu and Wu, Jialong and Feng, Ningya and Xiao, Chenjun and Li, Dong and Hao, Jianye and Wang, Jianmin and Long, Mingsheng},
  journal={arXiv preprint arXiv:2310.00344},
  year={2023}
}

@article{tassa2018deepmind,
  title={Deepmind control suite},
  author={Tassa, Yuval and Doron, Yotam and Muldal, Alistair and Erez, Tom and Li, Yazhe and Casas, Diego de Las and Budden, David and Abdolmaleki, Abbas and Merel, Josh and Lefrancq, Andrew and others},
  journal={arXiv preprint arXiv:1801.00690},
  year={2018}
}

@inproceedings{zhang2019solar,
  title={Solar: Deep structured representations for model-based reinforcement learning},
  author={Zhang, Marvin and Vikram, Sharad and Smith, Laura and Abbeel, Pieter and Johnson, Matthew and Levine, Sergey},
  booktitle={International conference on machine learning},
  pages={7444--7453},
  year={2019},
  organization={PMLR}
}

@article{haarnoja2018soft,
  title={Soft actor-critic algorithms and applications},
  author={Haarnoja, Tuomas and Zhou, Aurick and Hartikainen, Kristian and Tucker, George and Ha, Sehoon and Tan, Jie and Kumar, Vikash and Zhu, Henry and Gupta, Abhishek and Abbeel, Pieter and others},
  journal={arXiv preprint arXiv:1812.05905},
  year={2018}
}

@article{schulman2017proximal,
  title={Proximal policy optimization algorithms},
  author={Schulman, John and Wolski, Filip and Dhariwal, Prafulla and Radford, Alec and Klimov, Oleg},
  journal={arXiv preprint arXiv:1707.06347},
  year={2017}
}

@inproceedings{hafner2019learning,
  title={Learning latent dynamics for planning from pixels},
  author={Hafner, Danijar and Lillicrap, Timothy and Fischer, Ian and Villegas, Ruben and Ha, David and Lee, Honglak and Davidson, James},
  booktitle={International conference on machine learning},
  pages={2555--2565},
  year={2019},
  organization={PMLR}
}

@article{watter2015embed,
  title={Embed to control: A locally linear latent dynamics model for control from raw images},
  author={Watter, Manuel and Springenberg, Jost and Boedecker, Joschka and Riedmiller, Martin},
  journal={Advances in neural information processing systems},
  volume={28},
  year={2015}
}

@article{hafner2023mastering,
  title={Mastering diverse domains through world models, 2023},
  author={Hafner, Danijar and Pasukonis, Jurgis and Ba, Jimmy and Lillicrap, Timothy},
  journal={URL https://arxiv. org/abs/2301.04104},
  year={2023}
}

\end{document}